\documentclass{svproc}
\usepackage{graphicx}
\usepackage{amsmath}
\usepackage{amsfonts}
\usepackage{amssymb}
\usepackage{graphicx}
\graphicspath{ {./Graphs/} }
\usepackage{url}

\begin{document}
\mainmatter              % start of a contribution
\title{Explorations in Self-Supervised Learning: Dataset Composition Testing for Object Classification}
\titlerunning{Dataset Composition Testing for Object Classification}  % abbreviated title (for running head)
%------------!! Don't forget to define \inst
\author{Raynor Kirkson E. Chavez*\inst{1} \and Kyle Gabriel M. Reynoso*\inst{1}
Prospero C. Naval Jr.\inst{1} \and Carlo R. Raquel\inst{1}}
\authorrunning{Chavez et al.} % abbreviated author list (for running head)
\tocauthor{Raynor Kirkson E. Chavez, Kyle Gabriel M. Reynoso,
Prospero C. Naval Jr., Carlo R. Raquel}
\institute{
    University of the Philippines - Diliman, Quezon City, Philippines,\\
    \email{rechavez@up.edu.ph}
    \email{kmreynoso@up.edu.ph} \\
    \emph{*equal contribution}
}

\maketitle              % typeset the title of the contribution

% Your abstract
\begin{abstract}
This paper investigates the impact of sampling and pretraining using datasets with different image characteristics on the performance of self-supervised learning (SSL) models for object classification. To do this, we sample two apartment datasets from the Omnidata platform based on modality, luminosity, image size, and camera field of view and use them to pretrain a SimCLR model. The encodings generated from the pretrained model are then transferred to a supervised Resnet-50 model for object classification. Through A/B testing, we find that depth pretrained models are more effective on low resolution images, while RGB pretrained models perform better on higher resolution images. We also discover that increasing the luminosity of training images can improve the performance of models on low resolution images without negatively affecting their performance on higher resolution images.


\end{abstract}

% Do not use any of these
% A category with only the three required fields
%\category{H.4.m}{Information Systems}{Miscellaneous}
%\category{D.2}{Software}{Software Engineering}
%A category including the fourth, optional field follows...
%\category{D.2.8}{Software Engineering}{Metrics}[complexity measures,performance measures]
%\terms{Delphi theory}
%\keywords{Speaker Identification} % NOT required for Proceedings
%-----------------------------
%Section: Introduction
%-----------------------------
\section{Introduction}
Self-supervised learning (SSL) has garnered increasing attention in the field of machine learning due to its ability to train on large amounts of unlabeled data and achieve good performance on a wide range of tasks through transfer learning. Some examples of computer vision tasks that have been successfully tackled using SSL include image classification, object detection, cross-model retrieval, and instance segmentation \cite{han2021}. SSL methods can be classified into two categories based on their architecture: contrastive and non-contrastive. Recently, researchers have focused on scaling SSL algorithms to work with even larger datasets and improving their performance by optimizing the encoding layer architecture which maps data points to the feature space \cite{fang2022}. However, the performance of these methods is still highly dependent on the characteristics of the input data, particularly the variation of features in the dataset.

To address this issue, researchers have begun to study the performance of SSL algorithms across different datasets and to generate training sets with specific variations in the input data \cite{jing2019}. In addition to data augmentation techniques, some researchers have explored the idea of training multimodal models while removing certain modalities, with the goal of avoiding overfitting on the original dataset. This approach has shown promising results in some cases, but the degree of improvement can vary greatly depending on the characteristics of the dataset \cite{alayrac2020}. These findings suggest that there may be other factors, beyond the procedural details of the model itself, that influence the performance of SSL algorithms \cite{ma2022}. Further research is needed to understand these connections and identify potential ways to improve the performance of these methods.

This paper explored different variations of input datasets to improve the performance of a (baseline) model.

%-----------------------------
%Section: Methodology
%-----------------------------
\section{Methodology}
In this study, we propose a method for self-supervised learning that leverages pre-generated datasets from the Omnidata platform\cite{omnidata} as input for pretraining with the SimCLR model. We chose the SimCLR model for its computational efficiency, as it requires fewer operations than state-of-the-art SSL models\cite{simclr}, allowing us to work with limited hardware (1 NVIDIA V100 Core). The datasets we use consist of apartment images generated from 3D models of various rooms. After pretraining, the learned encodings are transferred to a supervised learning model for finetuning.
%---------Subsection: Datasets Used---------
\subsection{Datasets Used}
For pretraining, we used the frl\_apartment\_0 and allensville datasets from the Omnidata starter dataset. These datasets consist of point-randomized simulated camera captures of 3D apartment room models. We used 512 x 512 RGB and depth images for model training, and also included associated camera information such as angle, distance, obliqueness, and point of view to characterize the datasets.
%---------Subsection: Pretraining---------
\subsection{Pretraining}
The two apartment image variations are represented as classes and fed as input into the SimCLR model which uses contrastive loss to learn features by comparison across classes.

Seven models were trained using different dataset variations. From the frl\_apartment\_0 and allensville datasets, only 3000 images (25\% of the generated samples) were sampled for each variation due to hardware constraints. The SimCLR model runs for 50 epochs with a batch size of 32 for every training run. The checkpoint of each distinct model trained under a variation is saved for the finetuning step.

The first two runs were varied across their data domain: RGB and depth. The first model was trained on RGB data while the second was fed black-and-white encoded depth data corresponding to the same images. The next run tested the application of the Brightness (2X) augmentation on the RGB image taken as input.

The remaining runs tested variations in luminosity and field of view. To determine variations of these features, K-means clustering was performed on the one-dimensional numerical values denoting these qualities for three groups (k=3): high, medium, and low. Luminosity is defined as the mean pixel average luminosity of the generated RGB images while field of view (in radians) is indicated in the associated camera properties file of the RGB images. Four distinct models were trained based on the following clusters: low luminosity, high luminosity, low field of view and high field of view.

%---------Subsection: Transfer Learning---------
\subsection{Transfer Learning}
The seven resulting models were tested on the object classification downstream task. The encoding from pretraining checkpoints were transferred to a ResNet 50 model whose last layer is unfrozen to allow for supervised learning. The models were run for 100 epochs with a batch size of 256 on two standard labelled image datasets: CIFAR-10 and STL-10. The CIFAR-10 dataset\cite{cifar10} consists of 60000 32x32 RGB images classified into 10 classes while the STL-10 dataset\cite{stl10} is a modified version of the CIFAR-10 dataset which contains fewer classified samples at a higher resolution (96x96) but more unlabelled instances of the same classes. Fourteen finetuned models resulted from running the seven models on two different datasets. An additional 2 models were generated by downsampling the STL-10 dataset into 32x32 pixel images as baseline for performance across data modalities (RGB and depth) for low resolution pictures.

The top 1 train and test, and top 5 train accuracies of each epoch is recorded and plotted as a graph to compare model reliability across iterations. The convergence of these graphs toward their peak as well as their comparative accuracies were analyzed to validate the claim that changes in the distribution of datasets affect model performance. From the results, the researchers also hypothesized about underlying factors causing these fluctuations that may be explored by further studies.

%-----------------------------
%Section: Experimental Results
%-----------------------------
\section{Experimental Results}

%---------Subsection: Modality Tests---------
\subsection{Modality Tests}
\begin{figure}[h]
\centering
\includegraphics[width = .4\textwidth]{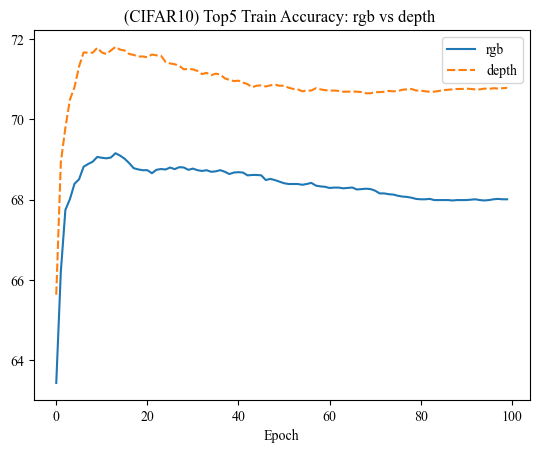}
\caption{Top 5 Training accuracy of the SimCLR model pretrained on 3000 RGB or depth images and finetuned and tested on the CIFAR10 dataset.}
\label{fig:float}
\end{figure}

\begin{figure}[h]
\centering
\includegraphics[width = .4\textwidth]{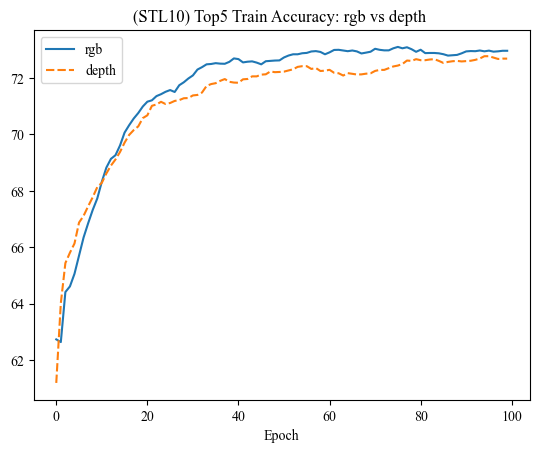}
\caption{Top 5 Training accuracy of the SimCLR model pretrained on 3000 RGB or depth images and finetuned and tested on the STL10 dataset.}
\label{fig:float}
\end{figure}

Our experimental results (Fig. 1 and Fig. 2) show that both SimCLR models pretrained on depth and RGB data perform well above average on the image classification downstream task at epoch 0, indicating that both models learned useful representations from the images during the pretraining step.

When evaluating the models on the CIFAR10 dataset at epoch 0, we observe that the depth model performs better overall compared to the RGB model. In contrast, when testing on the STL10 dataset at epoch 0, the performances of both models are very close, with the RGB model slightly outperforming the depth model.

These observations suggest that the SimCLR model trained on depth data may have learned more effective representations from the image data during the pretraining step compared to the SimCLR model trained on RGB data.

\begin{figure}[h]
\centering
\includegraphics[width = .4\textwidth]{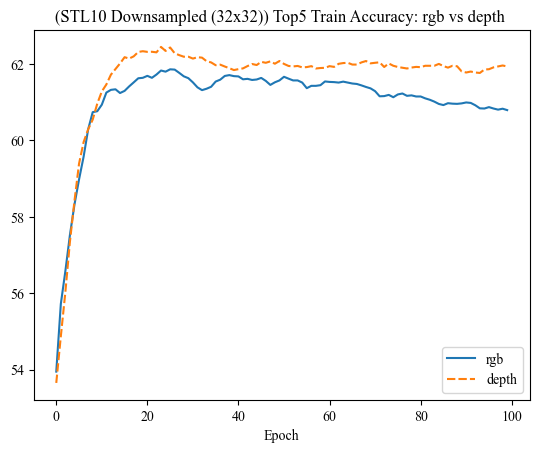}
\caption{Top 5 Training accuracy of the SimCLR model pretrained on 3000 RGB or depth images and finetuned and tested on the STL10 dataset.}
\label{fig:float}
\end{figure}

Our experimental results (Table 1) following finetuning show that the SimCLR model pretrained on RGB performed slightly better than the model trained on depth when tested on the STL10 dataset, while the opposite was true when testing on the CIFAR10 dataset, with the model pretrained on depth outperforming the one pretrained on RGB. This suggests that a SimCLR model pretrained on depth may be more effective on low resolution images, while models pretrained on RGB may excel on higher resolution or more detailed images.

The improved performance of both models on the STL10 dataset warrants further investigation, particularly given the differences in the number and size of the training images between the two datasets. One potential avenue for exploration is the impact of input image size, as the CIFAR10 dataset consists of 32x32 pixel images while the STL10 dataset has 96x96 pixel images.

Our subsequent analysis, in which we downsampled the STL10 dataset during the finetuning phase, supports this hypothesis, as it resulted in the depth pretrained model outperforming the RGB pretrained model, reversing the trend observed without downsampling (Table 2). These results provide evidence that the depth pretrained model may be more effective on low resolution images, while the RGB pretrained model performs better on higher resolution images.

To further investigate these findings, we conduct additional experiments that vary the brightness, image size, luminosity, and FOV of the pretraining dataset. Additionally, we may explore the performance of other self-supervised learning methods, such as MoCo, on these tasks in the future. These efforts will allow us to gain a deeper understanding of the factors that contribute to the relative performance of these models.

%---------Subsection: Brightness Tests---------
\subsection{Brightness Tests}

\begin{figure}[h]
\centering
\includegraphics[width = .4\textwidth]{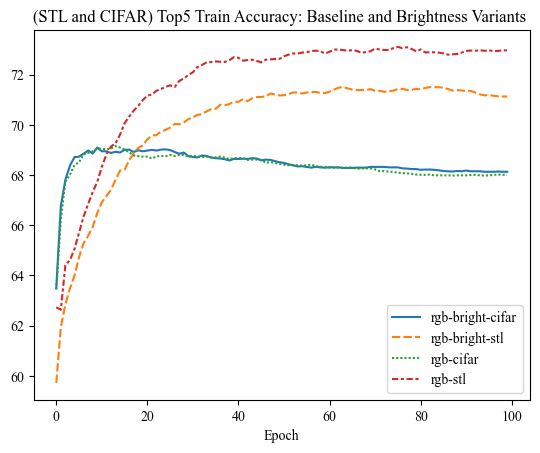}
\caption{Top 5 Training accuracy of the SimCLR baseline model and the high brightness variant finetuned and tested on the CIFAR10 and STL10 datasets.}
\label{fig:float}
\end{figure}

The performance of an RGB model finetuned on STL10 and adjusted to be two times brighter than normal was found to be lower than that of the baseline RGB model finetuned on STL10 (Fig. 4). However, finetuning the brightness-adjusted RGB model on CIFAR10 resulted in performance that was roughly equivalent to that of the baseline RGB model finetuned on CIFAR10.

The cause of this discrepancy is currently unknown, and further experimentation, such as repeating the test with a downsampled STL10 dataset on the brightness-adjusted pretrained RGB model, may be necessary to gain a deeper understanding of these results.

%---------Subsection: Luminosity Tests---------
\subsection{Luminosity Tests}

\begin{figure}[h]
\centering
\includegraphics[width = .4\textwidth]{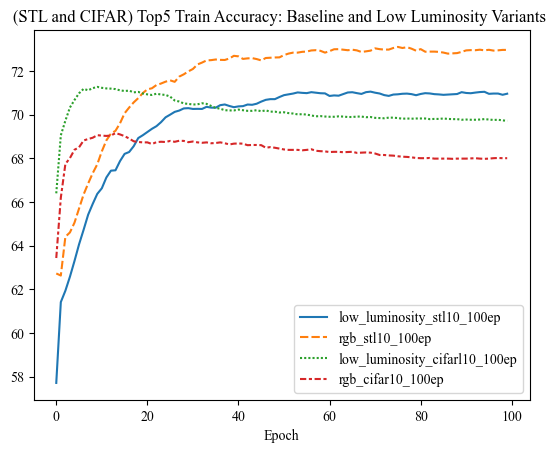}
\caption{Top 5 Training accuracy of the SimCLR baseline model and the low luminosity variant finetuned and tested on the CIFAR10 and STL10 datasets.}
\label{fig:float}
\end{figure}
\begin{figure}[h]
\centering
\includegraphics[width = .4\textwidth]{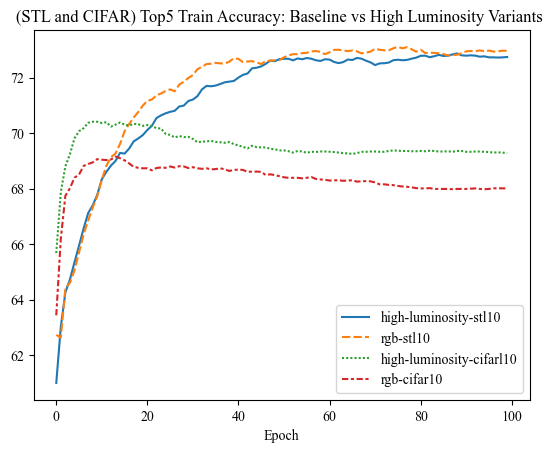}
\caption{Top 5 Training accuracy of the SimCLR baseline model and the high luminosity variant finetuned and tested on the CIFAR10 and STL10 datasets.}
\label{fig:float}
\end{figure}

The results show that the low-luminosity adjusted pretrained model outperforms the same model pretrained on stl10 when finetuned on cifar. In contrast, the baseline rgb models exhibit the opposite trend when finetuned on STL10 and CIFAR10 (Fig. 5 and Fig. 6).

Additionally, the high-luminosity pretrained variant performs similarly to the baseline model pretrained on rgb and finetuned on STL10. However, the high-luminosity model variant shows significant improvements in performance when finetuned on CIFAR10 compared to the baseline rgb model finetuned on CIFAR10. These findings suggest that increasing luminosity may enhance the performance of models on low resolution images without compromising performance on higher resolution images.

Although the baseline rgb model finetuned on STL10 remains the top performer among the four models, the low-luminosity adjusted model exhibits improved base performance when finetuned on CIFAR10, but decreased performance when finetuned on STL10. The reason for this discrepancy in performance is currently unknown.

Additionally, it is worth noting that the improvement in performance of the low-luminosity adjusted model when finetuned on CIFAR10 may be due to the increased contrast in the images, as low-luminosity images tend to have higher contrast compared to high-luminosity images. However, the negative impact on performance when finetuning on STL10 is more difficult to explain and warrants further investigation. It is possible that the increased contrast in the low-luminosity images may interfere with certain features that are important for the stl10 dataset.

Overall, these findings suggest that the impact of luminosity on the performance of deep learning models is complex and dependent on the specific characteristics of the dataset. Further research is necessary to fully understand the mechanisms behind these trends and how to effectively leverage luminosity to improve model performance.

%---------Subsection: Field of View Tests---------
\subsection{Field of View Tests}

\begin{figure}[h]
\centering
\includegraphics[width = .4\textwidth]{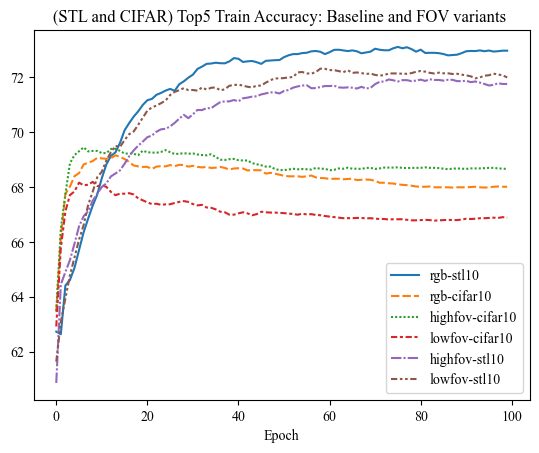}
\caption{Top 5 Training accuracy of the SimCLR baseline model and the high and low FOV variants finetuned and tested on the CIFAR10 and STL10 datasets.}
\label{fig:float}
\end{figure}

The experimental results for field of view variance demonstrate that training on images with varying field of views did not significantly impact performance on the STL10 dataset, but did result in marginally better performance on the CIFAR10 dataset for the high FOV group compared to the low FOV group and the baseline RGB model (Fig. 7).

Field of view denotes how wide the camera angle is, hence, it can be used as a proxy for the number of features represented in the training data, with higher FOV corresponding to a wider view and potentially more representative datasets as indicated by the models finetuned on the STL10 dataset. However, the models pretrained on the CIFAR10 dataset showed a different trend, potentially indicating the need for a different encoding of the data for effective classification. It is worth noting that the CIFAR10 dataset consists of 32x32 pixel images, while the STL10 dataset has 96x96 pixel images, which may account for the differing results. This suggests that the benefits of high FOV may not generalize to vision tasks with smaller input dimensions.

\begin{table}[]
\begin{center}
\begin{tabular}{|ll|l|}
\hline
\textbf{}                    & \textbf{STL10}   & \textbf{CIFAR}   \\ \hline
\textit{\textbf{RGB}}        & 73.11 @ Epoch 75 & 69.15 @ Epoch 13 \\
\textit{\textbf{Depth}}      & 72.78 @ Epoch 94 & 71.81 @ Epoch 13 \\ \hline
\textit{\textbf{Difference}} & 0.33             & -2.65            \\ \hline
\end{tabular}
\end{center}
\caption{Maximum Top5 Accuracies of the models finetuned and Tested on CIFAR10 and STL10}
\end{table}

%-----------------------------
%Section: Conclusion
%-----------------------------
\section{Conclusion}
In summary, the results of our experiments show that both SimCLR models pretrained on depth and RGB data perform well on the image classification downstream task, with the depth model performing slightly better overall on the CIFAR10 dataset and the RGB model performing slightly better on the STL10 dataset at epoch 0. However, after finetuning, the performance of the models on the two datasets is reversed, with the SimCLR model pretrained on depth performing better on the CIFAR10 dataset and the model pretrained on RGB performing better on the STL10 dataset. This suggests that the depth pretrained model may be more effective on low resolution images, while the RGB pretrained model performs better on higher resolution images. The results of our analysis, in which we downsampled the STL10 dataset during finetuning, support this hypothesis. 

In addition, we found that increasing the luminosity of the images used for training may enhance the performance of models on low resolution images without compromising performance on higher resolution images. Finally, we found that the field of view of the images used for training did not significantly impact performance on the STL10 dataset, but did result in slightly better performance on the CIFAR10 dataset for the high FOV group compared to the low FOV group and the baseline RGB model. The benefits of high FOV may not generalize to vision tasks with smaller input dimensions. Further research is necessary to fully understand the mechanisms behind these trends and how to effectively leverage these factors to improve model performance.

\begin{table}[]
\begin{center}
\begin{tabular}{|ll|l|}
\hline
\textbf{Base Model}     & \textbf{Finetuned}  & \textbf{Max. Top5 Acc.} \\ \hline
\textit{\textbf{rgb}}   & CIFAR10             & 69.15 @ Epoch 13            \\
\textit{\textbf{depth}} & CIFAR10             & 71.80 @ Epoch 13            \\
\textit{\textbf{rgb}}   & STL10               & 73.10 @ Epoch 75            \\
depth                   & STL10               & 72.78 @ Epoch 94            \\
rgb                     & STL10 (32x32) & 61.86 @ Epoch 25            \\
depth                   & STL10 (32x32) & 62.44 @ Epoch 23            \\
rgb (2x bright)         & STL10               & 71.51 @ Epoch 63            \\
rgb (2x bright)         & CIFAR10             & 69.09 @ Epoch 9             \\
rgb (low lum.)          & STL10               & 71.06 @ Epoch 69            \\
rgb (low lum.)          & CIFAR10             & 71.27 @ Epoch 9             \\
rgb (high lum.)         & STL10               & 72.87 @ Epoch 88            \\
rgb (high lum.)         & CIFAR10             & 70.41 @ Epoch 8             \\
rgb (low FOV)           & STL10               & 72.31 @ Epoch 58            \\
rgb (low FOV)           & CIFAR10             & 68.19 @ Epoch 8             \\
rgb (high FOV)          & STL10               & 71.93 @ Epoch 82            \\
rgb (high FOV)          & CIFAR10             & 69.44 @ Epoch 6             \\ \hline
\end{tabular}
\end{center}
\caption{Max. Top 5 Performance of the base model variants and corresponding finetuning datasets on the image classification downstream task}
\end{table}

\section{Acknowledgements}
Authors Chavez and Reynoso contributed equally to this paper. The researchers would like to express their gratitude to their advisors, Dr. Prospero C. Naval Jr. and Carlo R. Raquel, for their valuable guidance, support, and advice throughout the development of this project. Thanks are also extended to members of the Computer Vision and Machine Intelligence Group (CVMIG) for their continuous feedback on the research work, which enriched the exploration of different avenues.

\nocite{*}
\bibliographystyle{IEEE}

\begin{thebibliography}{1}  
\bibitem{han2021}Han, X., Zhang, Z., Ding, N., Gu, Y., Liu, X., Huo, Y., Qiu, J., Yao, Y., Zhang, A., Zhang, L., Han, W., Huang, M., Jin, Q., Lan, Y., Liu, Y., Liu, Z., Lu, Z., Qiu, X., Song, R., Tang, J., Wen, J., Yuan, J., Zhao, W. \& Zhu, J. Pre-trained models: Past, present and future. {\em AI Open}. \textbf{2} pp. 225-250 (2021), https://doi.org/10.1016/j.aiopen.2021.08.002

\bibitem{fang2022}Fang, P., Li, X., Yan, Y., Zhang, S., Kang, Q., Li, X. \& Lan, Z. Connecting the Dots in Self-Supervised Learning: A Brief Survey for Beginners. {\em Journal Of Computer Science And Technology}. \textbf{37}, 507-526 (2022,6), https://doi.org/10.1007/s11390-022-2158-x

\bibitem{jing2019}Jing, L. \& Tian, Y. Self-Supervised Visual Feature Learning With Deep Neural Networks: A Survey. {\em IEEE Transactions On Pattern Analysis And Machine Intelligence}. \textbf{43} pp. 4037-4058 (2019)

\bibitem{alayrac2020}Alayrac, J., Recasens, A., Schneider, R., Arandjelović, R., Ramapuram, J., De Fauw, J., Smaira, L., Dieleman, S. \& Zisserman, A. Self-Supervised MultiModal Versatile Networks. {\em Advances In Neural Information Processing Systems}. \textbf{33} pp. 25-37 (2020), https://doi.org/10.48550/arXiv.2006.16228

\bibitem{ma2022}Ma, M., Ren, J., Zhao, L., Testuggine, D. \& Peng, X. Are Multimodal Transformers Robust to Missing Modality?. {\em 2022 IEEE/CVF Conference On Computer Vision And Pattern Recognition (CVPR)}. pp. 18156-18165 (2022)

\bibitem{omnidata}Eftekhar, A., Sax, A., Malik, J. \& Zamir, A. Omnidata: A Scalable Pipeline for Making Multi-Task Mid-Level Vision Datasets From 3D Scans. {\em Proceedings Of The IEEE/CVF International Conference On Computer Vision (ICCV)}. pp. 10786-10796 (2021,10)

\bibitem{simclr}Chen, T., Kornblith, S., Norouzi, M. \& Hinton, G. A Simple Framework for Contrastive Learning of Visual Representations. {\em Proceedings Of The 37th International Conference On Machine Learning}. (2020)

\bibitem{stl10}Coates, A., Ng, A. \& Lee, H. An Analysis of Single-Layer Networks in Unsupervised Feature Learning. {\em Proceedings Of The Fourteenth International Conference On Artificial Intelligence And Statistics}. \textbf{15} pp. 215-223 (2011,4,11), https://proceedings.mlr.press/v15/coates11a.html

\bibitem{cifar10}Krizhevsky, A. Learning Multiple Layers of Features from Tiny Images. {\em University Of Toronto}. (2012,5)

\end{thebibliography}

\end{document}